# Physics and semantic informed multi-sensor calibration via optimization theory and self-supervised learning

**Shmuel Y. Hayoun[*], Meir Halachmi[*], Doron Serebro, Kfir Twizer, Elinor Medezinski, Liron Korkidi, Moshik Cohen and Itai Orr[†]**

**Wisense Technologies Ltd., Israel**

## Abstract

Achieving safe and reliable autonomous driving relies greatly on the ability to achieve an accurate and robust perception system; however, this cannot be fully realized without precisely calibrated sensors. Environmental and operational conditions as well as improper maintenance can produce calibration errors inhibiting sensor fusion and, consequently, degrading the perception performance. Traditionally, sensor calibration is performed in a controlled environment with one or more known targets. Such a procedure can only be carried out in between drives and requires manual operation; a tedious task if needed to be conducted on a regular basis. This sparked a recent interest in online targetless methods, capable of yielding a set of geometric transformations based on perceived environmental features, however, the required redundancy in sensing modalities makes this task even more challenging, as the features captured by each modality and their distinctiveness may vary. We present a holistic approach to performing joint calibration of a camera-lidar-radar trio. Leveraging prior knowledge and physical properties of these sensing modalities together with semantic information, we propose two targetless calibration methods within a cost minimization framework once via direct online optimization, and second via self-supervised learning (SSL).

## 1. Introduction

With the increasing level of vehicle autonomy in recent years the idea of sensor fusion to improve perception capabilities[1–6] has gained much attention and is currently a major research topic. Understandably, the effectiveness of the fusion process is correlated to the level of correspondence between multiple measurements of the same object. Therefore, to maintain accurate perception over time, one must retain precisely calibrated sensors. This includes both the intrinsic calibration of each sensor as well as the extrinsic calibrations between all sensors. To illustrate the importance of sensor calibration, we show in Fig. 1, a sample frame from our dataset with the measurements from the camera, lidar and radar projected onto each other coordinates. Fig. 1a and Fig. 1b show the front view and the top view of the sensor measurements respectively. Each view is split along the middle, where the uncalibrated state is displayed in the left half and the calibrated state, resulting from our proposed approach, in the right half. In Fig. 1a the uncalibrated and calibrated states of the lidar and radar with respect to the camera are shown; on the left there is a clear misalignment between the projected lidar points (colored points) and the scene, as well as between the radar dynamic track detections (white '+' markers) and the moving vehicles. On the right-hand side, the sensors are well calibrated using our method. In Fig. 1b a bird's-eye view of the uncalibrated and calibrated lidar (colored points) and radar (black points) is given with x and y axes in meters; on the right the alignment correctness is discernible from the overlapping detections of objects such as vehicles and the right guard rail, unlike the severe misalignment seen on the left.

[*] Equal contribution

[†] Corresponding author: itaiorr@gmail.com

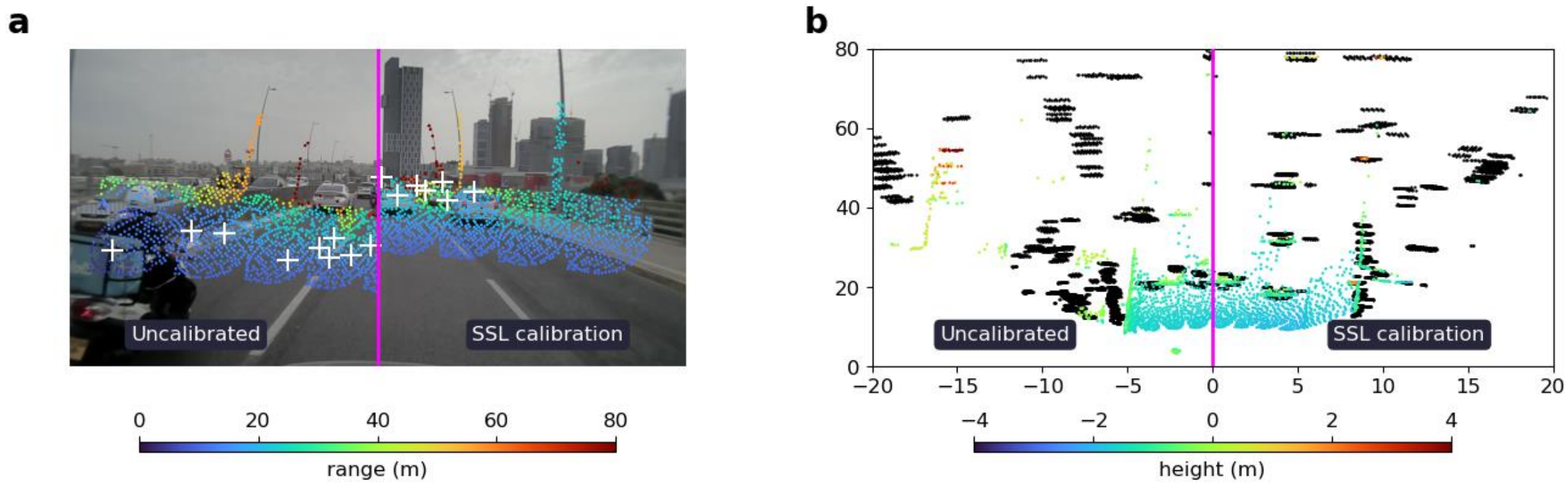

**Fig. 1 | Physics and semantic informed calibration.** Front view (**a**) and top view (**b**) of the sensor spatial measurements. Each view is split along the middle, where the uncalibrated state is displayed in the left half and the calibrated state, resulting from our proposed approach, in the right half. In (**a**) the uncalibrated and calibrated states of the lidar and radar with respect to the camera are shown; on the left there is a clear misalignment between the projected lidar points (coloured points) and the scene, as well as between the radar dynamic track detections (white '+' markers) and the moving vehicles. On the right-hand side, the sensors are well calibrated using our method. In (**b**) a bird's-eye view of the uncalibrated and calibrated lidar (coloured points) and radar (black points) is given with x and y axes in meters; on the right the alignment correctness is discernible from the overlapping detections of objects such as vehicles and the right guard rail, unlike the severe misalignment seen on the left.

Current calibration methods are usually categories as either target-based or target-less. Target-based calibration is performed in a controlled environment. i.e., certain perceivable objects are known (pose, shape, and size) and the scene is usually static. The specific targets are used to obtain simultaneous measurements from multiple sensors that can be registered using geometric alignment. The result is the calibrated extrinsic and/or intrinsic parameters. The controlled targets are designed according to the intended sensor types in such a way that target measurements are clear and thus registration is more precise. e.g., planar surfaces with a distinct pattern (e.g., checkered) for cameras[7–10], planar surfaces for lidars[7–14] and trihedral corner reflectors for radars[12–15].

Although the manner and environment in which they are conducted makes target-based methods highly accurate, they are also the reason why such techniques are not practical for every-day use. Sensors mounted to drivable platforms are subject to vibrations and weather changes, can be mishandled (e.g., improper installment or calibration) or suffer impacts. Having their recalibration done manually by an expert using dedicated equipment is simply not a viable or scalable solution, and it is therefore paramount that autonomous vehicles possess the ability to recover from an uncalibrated state online and automatically.

Targetless methods are inherently suited for online use. Instead of using specialized apparatuses, the calibration is conducted based on perceived environmental features from natural scenes. For autonomous vehicles this would mean urban roads and highways. As such, the calibrated state is a state of maximal cross-sensor alignment of a set of some distinctive matching features. This type of calibration can therefore be regarded as two subsequent problems. The first is meaningful data extraction and interpretation, and the second is finding the best alignment based on those features.

When calibrating multiple instances of the same sensing modality (e.g., multiple cameras), cross-sensor alignment can usually be defined straightforwardly in terms of the correspondence between the sensor measurements (e.g., pixelwise correspondence between images in terms of RGB values, edges, etc.). However, when calibrating different sensing modalities such correspondences may not be so simple to define, since the features captured by each modality and their distinctiveness may vary. This makes targetless calibration a challenging task and a generally less accurate approach compared to the target-based methodology.

Existing approaches, pertaining almost always to extrinsic calibration, are usually based on the use of geometric features, odometry and object tracking, physical properties and behavior of different modalities, and deep-learning-based semantics matching and regression.

Geometric approaches usually rely on matching distinctive spatial features between the different modalities. Various methods suggested the use of distinct line features and edges alignment to extrinsically calibrate a camera and lidar[16–19].

The calibration task was formulated as a registration problem between multiple planar regions, which are visible to both sensors and demonstrated how the camera intrinsic calibration can be included[19]. However, the solution required that the regions be associated a priori.

Following a different approach, after applying structure from motion on a series of consecutive images, 3D registration of a lidar point cloud and a camera-generated point cloud resulted in the camera-lidar extrinsic calibration[20]. A method for calibrating multiple lidars and radars in an automotive setup used the lidars to construct a 3D reference map of the environment, to which the radar detections were then registered, thereby obtaining the relative sensor poses[21].

Odometry-based methods use ego-motion estimations for each sensor to extrapolate the rigid body transformations between them. Such approaches were applied to various sensor setups such as camera and lidar[22], stereo cameras and lidar[23] and camera and radar[24]. Object tracking has enabled sensor calibration by aligning one or more track traces to discern the relative poses between the sensors[25].

Physical attributes of different modality signals have also been utilized to define criteria for measurement correspondence. The predominant approach for physics-informed, camera-lidar calibration is the correlation between lidar reflectance and image intensity[22,26–28]. Another technique performs the calibration by matching the estimated normals at the lidar points to the image intensity[29].

Deep neural networks (DNNs) have been used in various ways for extrinsically calibrating sensors of different types. Early approaches employed DNNs to assist in extracting features for use within an optimization network. Some used semantic information from camera and lidar to define a mutual-information-based objective function[30,31]. Others used a DNN to segment certain semantics in the image that could either be heuristically matched to detections from another sensor[32,33] or to filter out certain objects[20]. RegNet[34], CalibNet[35], CalibRCNN[36], RGGNet[37] are examples of DNNs used for end-to-end calibration of a camera-lidar setup. Whether explicitly or implicitly, the feature extraction and matching are incorporated into the networks and the calibration parameters are regressed. Using a slightly different approach with respect to the output, LCCNet [38], CFNet[39] and DXQ-Net[40] were trained to output a measure of the transformation correction needed for the given input. LCCNet did this directly, whereas CFNet and DXQ-Net provided this in a pixelwise manner over the projected lidar point cloud, termed calibration flow.

Ultimately, all these methods relied on some form of ground-truth data for training. In terms of applicability, this constitutes a significant drawback. Conversely, self-supervised learning does not require any data labeling, making it especially suitable in situations where ground-truth annotations are not readily available and too costly to produce.

The use of self-supervision has gained popularity in a growing number of fields. Self-supervised learning is a training method where one part of the signal is used to predict another part of the signal, thereby exposing underlying representations within the data. For example, this method was used for super-resolving a radar array[42], up-sampling a camera frame[43–46] or densification of lidar measurements[47–49].

Recent studies aimed to reformulate various problems which were previously solved using supervised learning to enable the use of self-supervision instead. A prominent example of this are recent solutions for generating image semantics, such as object detection[50–52] and semantic segmentation[53–55]. Self-supervision has also been used in different applications involving sensor measurement registration[56–58]. It has also been applied to intrinsic calibration of a monocular camera using information from subsequent images[59].

The targetless solutions mentioned so far perform calibration between two sensing modalities only. When considering the desired redundancy by parts of the autonomous vehicles industry, these approaches fall short of a more comprehensive solution, capable of handling a camera-lidar-radar setup calibration.

Previous work on methods for calibrating the camera-lidar-radar trio is scarce. An online, targetless methodology for extrinsic multi-sensor calibration was outlined and demonstrated on the trio setup[60]. Track-to-track association between sensor pairs was used for both detecting cross-sensor miscalibration and performing recalibration when necessary. One sensor was selected as a common reference, to which all other sensors were calibrated. The transformation between any two sensors could then be derived from their respective transformations to the reference sensor. This methodology does not take advantage of relevant cross-sensor correspondences making it more vulnerable to different environmental conditions. Meaning, choosing the camera as the reference sensor would cause the entire calibration process to fail under poor lighting. Furthermore, the reliance on dynamic object tracks exclusively means that this approach cannot be applied unless at least one such object exists and is being tracked by all the sensors.

In this work we propose a joint multi-sensor calibration approach which consists of deducing the correct relative alignments between the sensors from several features across their respective signal measurements. The features are extracted and based on known physical properties of the different sensing modalities and independently generated semantics.

We propose two different methods to the calibration problem. The first consists of solving an optimization problem under a set of pairwise and global self-consistency constraints. The second approach is to train a DNN in a self-supervised manner without any ground truth, on the same minimization problem by using the objective function as the training loss. Once the DNN is trained, it is capable of single frame operation, making it robust to environmental and operational conditions.

# 2. Results

## 2.1. Dataset

To evaluate our approaches, we used a sensor setup which included a camera, lidar and radar. The data was collected in different, uncontrolled urban and highway environments over multiple days where the sensors setup was mounted at the beginning of each day and removed at the end of each day. Meaning, sensor alignment varied across locations and days. The dataset contained 240,000 frames, where each frame contains temporally synced measurements from the camera, lidar and radar. The data was split to 200,000 for training and 40,000 for validation. The data for the validation was taken from different days and locations to avoid the appearance of similar frames in both datasets, which could have occurred in the case of simple random split. An example of a frame from the dataset collected is provided in Fig. 1.

The sensor setup utilized a camera with a field of view (FOV) of $116°\text{H} \times 58.5°\text{V}$ that captures $1824 \times 940$ resolution images, a lidar with a FOV of $60°\text{H} \times 24°\text{V}$ and $0.25°\text{H} \times 0.25°\text{V}$. The radar used was a frequency-modulated continuous wave (FMCW), collocated multiple-in multiple-out (MIMO) radar with a 79GHz carrier frequency. The angular resolution of the radar was $1.6°\text{H} \times 6°\text{V}$ providing measurements in 4 dimensions: azimuth, elevation, range, and Doppler. An FMCW radar transmits a linear chirp signal whose frequency increases linearly with time. By means of signal processing, mainly Fast Fourier Transform, it is possible to extract useful information from the raw signal, such as range, velocity, and direction of arrival. MIMO radar is composed of multiple transmitter (Tx) and receiver (Rx) antennas. Each transmitter can transmit a waveform independently of the other transmitting antennas, whereas each of the receiving antennas can also receive these signals independently. By processing measurements from different transmitting and receiving antennas, one can create a virtual aperture whose size is larger than the physical one. Meaning, an antenna array composed of $NTx$ transmitters and an array of $NRx$ receivers results in a virtual array of $NTx \times NRx$ channels. The radar FOV was configured to match the lidar's FOV. Both the lidar and the radar were configured to minimum range of 5m and maximum range of 80m.

An additional dataset for quantitative comparison was collected under controlled, stationary conditions. The controlled setup included three trihedral reflective corners placed in a leveled empty lot at varying locations to allow for spatially diverse error measurement. These targets were recorded for 150 frames by each of the sensors and their positions were manually pinpointed within each modality's coordinate system. These recordings took place in adjacent to the data collection for the validation set, thus ensuring similar calibration conditions.
An example frame from the controlled test setup is provided in Fig. 2. A view of the controlled test setup from the camera's perspective is shown in Fig. 2a. Calibration errors of the camera-paired transformations were extracted from images in terms of horizontal and vertical pixelwise deviations ($\delta u$ and $\delta v$, respectively). These were then converted to 2D angular errors of azimuth and elevation. A top view illustration of the controlled test setup is shown in Fig. 2b. The lidar-radar calibration and the camera-lidar-radar consistency errors were measured in 3D coordinates expressed in azimuth ($\delta\theta_z$), elevation ($\delta\theta_x$), and range ($\delta r$).

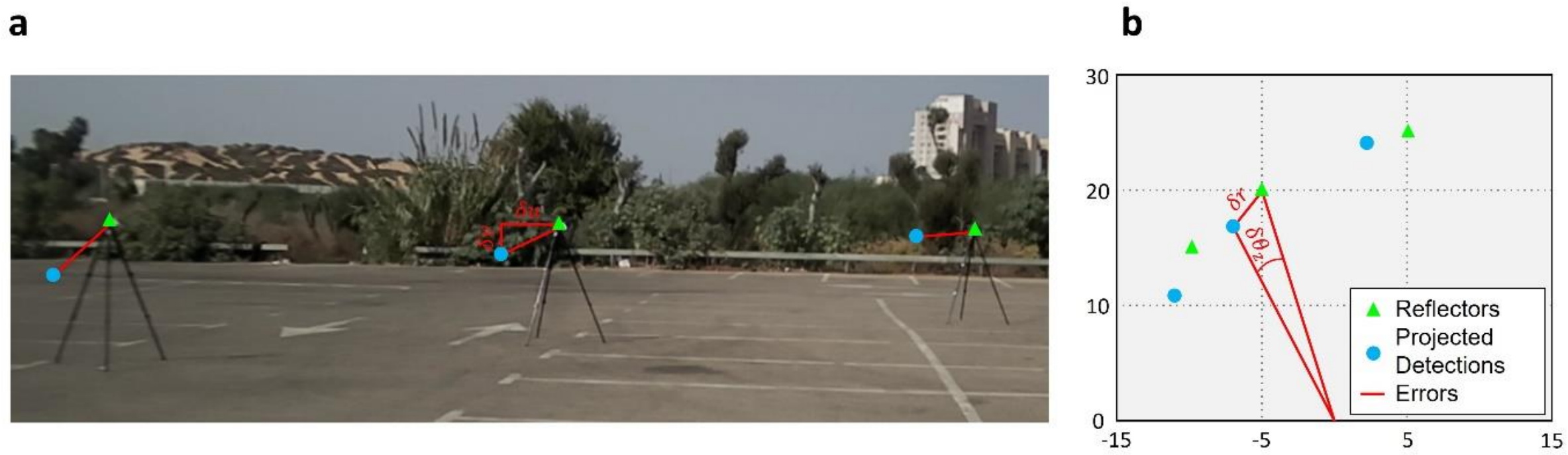

**Fig. 2 | Controlled test setup.** The controlled test setup was made up of three trihedral corner reflectors (green triangles) at varying range-azimuth combinations to allow for spatially diverse error measurement. Simulated projections (blue circles) were added to illustrate error measurement (red lines) expressed in azimuth ($\delta\theta_z$), elevation ($\delta\theta_x$), and range ($\delta r$). **a**, controlled test setup from the camera's perspective, where initially the angular errors were measured pixelwise horizontally ($\delta u$) and vertically ($\delta v$). **b**, top view of the controlled test setup; x and y axis are in meters.

## 2.2. Experiments

Qualitative assessment of our proposed methods can be done by projecting one sensor's measurements onto another sensor's frame of reference. Sample results from the validation dataset for both the optimization-based and SSL-based configurations are provided in Fig. 3, where the results show good correspondence between the sensor measurements. From left to right, the camera-lidar alignment is shown in the image plane where the colormap used for the lidar point cloud represents the range in meters. The top view plot displays the lidar-radar correspondence, where the colormap is used to represent the lidar points' height in meters, whereas the radar detections are displayed in black to better distinguish between the two sensing modalities. To illustrate the camera-radar calibration, we used the centers of mass from tracked radar clusters, as it makes it easier for the reader to visualize the association of the dynamic objects in the camera-radar pair.

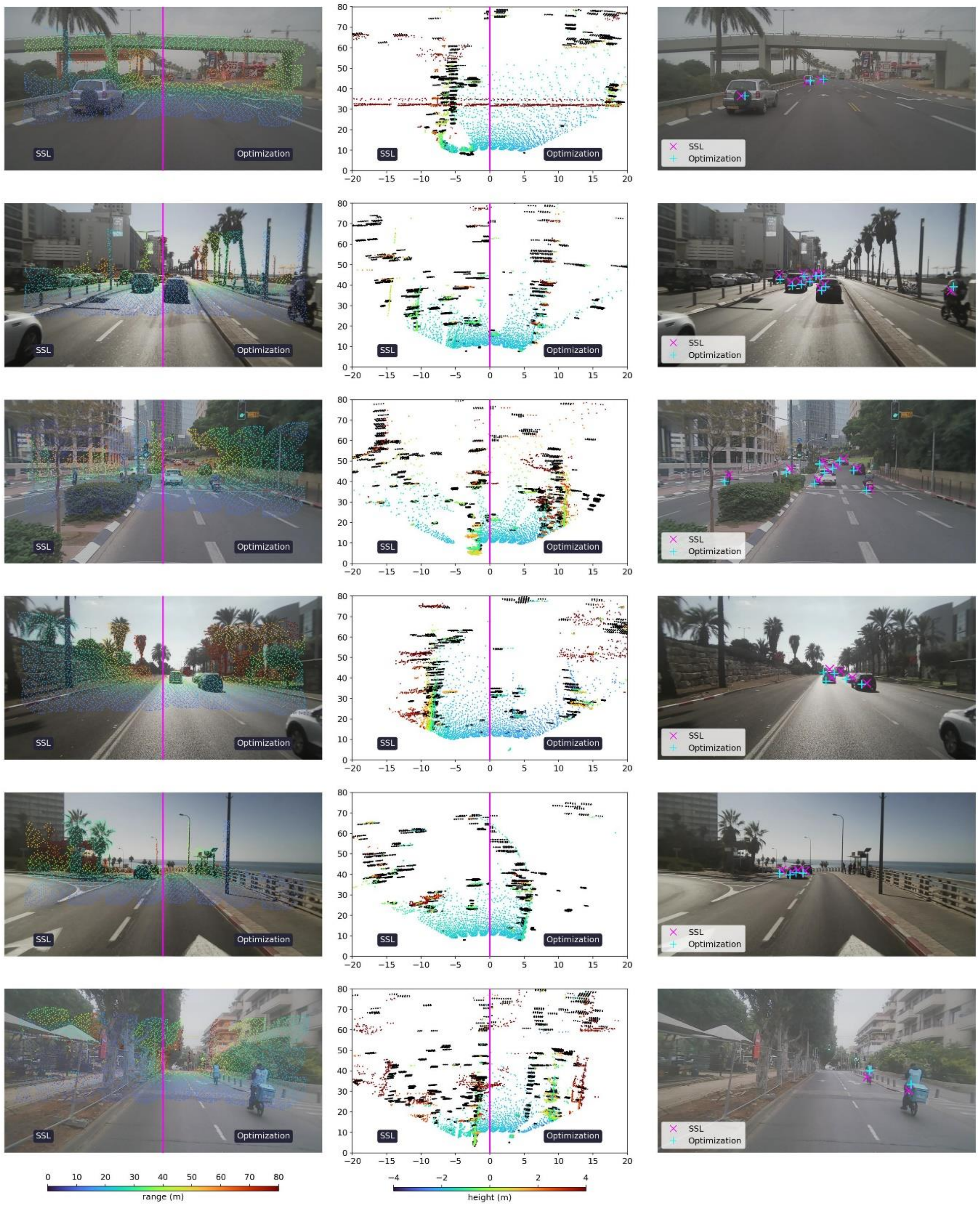

**Fig. 3 | Sample results from the validation dataset.** From left to right: Calibration results of the optimization-based and SSL-based methods for camera-lidar, lidar-radar and camera-radar, respectively. On the left column, the camera-lidar calibration showing the projected lidar point cloud onto its corresponding image, with colormap representing range in meters. In the middle column, the lidar-radar calibration is portrayed in bird's eye view. The lidar is represented in a colormap for height in meters, whereas the radar detections are represented as black points. For both pairs (camera-lidar and lidar-radar), each frame is split along the middle, where the optimization-based calibration is displayed in the right half and the SSL-based calibration in the left half. On the right column, center of mass from tracked radar clusters are projected onto the corresponding images using the calibrations from both proposed methods.

Quantitative evaluations were carried out using the controlled test setup. The methodology consisted of deriving all three pairwise calibrations from the validation dataset, then applying the different transformations matrices to the controlled test measurements and calculating the resulting alignment errors.

As a baseline, we compared our proposed optimization-based and SSL-based methods to the object-tracking-based multi-sensor calibration[60]. Due to lack of available sources, we used our implementation of this method and replaced the original camera-based 3D object detection[61] with a stronger detector[62]. The errors reported in Table 1 are an average over the validation dataset and all three targets in the controlled test setup.

| | | **Pairwise calibration errors** | | | | | | | | | **Global calibration errors** | | |
|---|---|---|---|---|---|---|---|---|---|---|---|---|---|
| | | **Camera-Lidar** | | | **Camera-Radar** | | | **Lidar-Radar** | | | **Camera-Lidar-Radar** | | |
| | | **Az [deg]** | **El [deg]** | **R [m]** | **Az [deg]** | **El [deg]** | **R [m]** | **Az [deg]** | **El [deg]** | **R [m]** | **Az [deg]** | **El [deg]** | **R [m]** |
| **Method** | Object-tracking-based calibration[60] | 0.55 | 2.01 | - | 1.92 | 3.13 | - | 1.46 | 1.87 | 0.22 | - | - | - |
| | Pairwise optimization | 0.21 | 0.28 | | 0.20 | 0.44 | | 0.97 | 0.21 | 0.14 | 1.43 | 0.83 | 0.01 |
| | Pairwise SSL | 0.07 | 0.09 | | 1.36 | 0.29 | | 1.17 | 1.09 | 0.30 | 0.14 | 0.61 | 0.01 |
| | Joint optimization | 0.22 | 0.27 | | 0.20 | 0.45 | | 0.36 | 0.72 | 0.14 | 0.09 | 0.01 | 0.01 |
| | Joint SSL | 0.08 | 0.20 | | 0.02 | 0.14 | | 0.51 | 0.61 | 0.13 | 0.02 | 0.02 | 0.01 |

**Table 1 | Quantitative evaluation of the proposed methods.** Average calibration errors, computed with respect to three reflective corners, are provided for all modality pairs as well as for the closed transformation loop between all sensors. Our optimization method was run once without the self-consistency loss in a decoupled manner (pairwise optimization) and once with the self-consistency loss in a coupled manner (joint three-way optimization). The SSL method was implemented once as three decoupled networks (pairwise SSL), and as a single network trained with the self-consistency loss.

In addition, a quantitative evaluation was also carried out for a closed-loop transformation to gauge the global self-consistency. The lidar was chosen as the reference sensor as it facilitated a 3D error representation. Accordingly, the closed-loop transform was the result of a composition of all pairwise transformations in a cyclic order beginning and ending with the lidar reference frame.

To evaluate the significance of including self-consistency considerations as part of the solution, we examined our proposed optimization-based and SSL-based methods with and without the self-consistency constraint. Under the non-consistent optimization-based configuration the optimization process was applied to each sensor pair separately using its corresponding objective function, yielding three independents pairwise calibrations. The pairwise SSL configuration was carried out by training three separate DNNs, each with a different pairwise loss function, to regress the corresponding sensors' calibration parameters solely from their respective measurements. The joint optimization consisted of performing the pairwise optimization which was then refined by a simultaneous joint optimization of all three sensors with respect to all pairwise objectives as well as a self-consistency constraint.

Meanwhile, the joint SSL configuration included a single DNN trained to predict all three sensors' pairwise calibrations given their measurements with the loss function being a composition of the pairwise losses and the self-consistency requirement. Additional details concerning the objective and loss functions can be found in the Methods section. The performances of each of these configurations are also reported in Table 1 highlighting the advantage of including the self-consistency constraints.

Additionally, we examined the joint SSL-based method's capacity for calibration under abrupt and large changes. This examination simulates real-world conditions where a sensor might get miss-aligned for various reasons, such as vibrations, impact, material fatigue and more. In each scene the sensors were given arbitrary initial alignment errors and the DNN was used to infer the correct alignments from that scene alone.

Qualitative results of this evaluation are shown in Fig. 4. The camera-paired uncalibrated and calibrated states are shown on the left with the projection of the lidar and radar measurements onto the corresponding image. On the right, a bird's-eye-view of the lidar-radar uncalibrated and calibrated states in each scene is shown in the lidar's coordinate frame. The results show good correspondence when applying our proposed SSL-based method and its robustness and generalization ability.

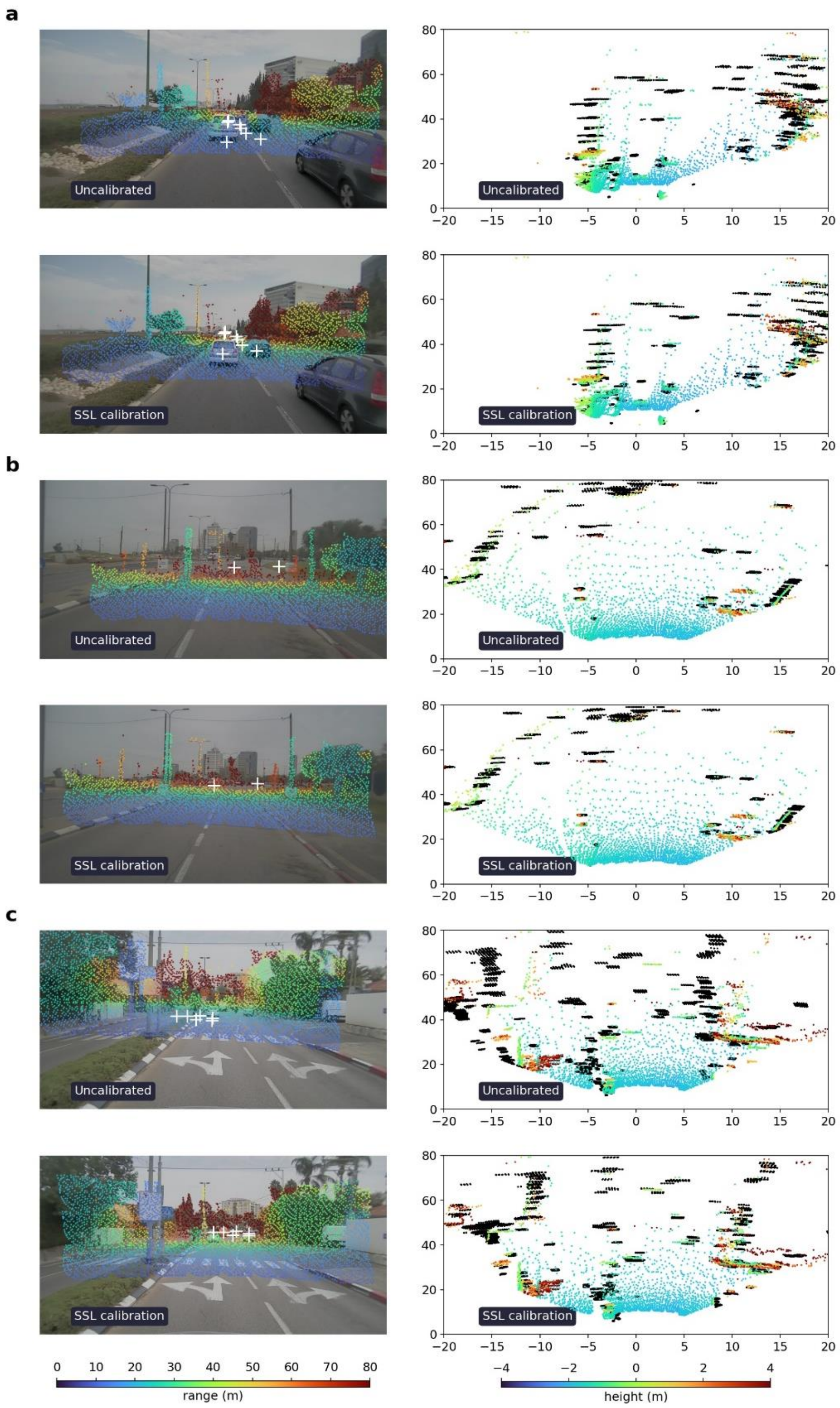

**Fig. 4 | Robustness to abrupt changes. a-c,** Examples of the joint SSL-based method in different scenes and various initial un-calibrated conditions. Sensor spatial measurements; x and y axes in meters. Each sample frame shows the uncalibrated on the top row and SSL-based calibrated on the bottom row. On the left column, lidar point cloud is projected on top of a camera image with colormap representing range in meters. In addition, center of mass of tracked radar clusters are shown using white '+' markers. On the right column, top view of the lidar point cloud with colormap representing height in meters. Radar detections are shown as black points.

We also examined the sensitivity of the proposed optimization-based and SSL-based methods with respect to different driving environments, mainly urban and highway scenarios. The results are provided in extended data figure 1 and extended data figure 2 showing advantage for the SSL-based method over the optimization-based method. Furthermore, the optimization-based method performs slightly better in urban conditions than in highway conditions, whereas the SSL-based method shows similar qualitative performance in both environments.

A quantitative examination of the two driving environments is provided in extended data table 1 confirming the qualitative results. There we see that the optimization-based method performs better in urban environments than in highway environments. In contrast, the SSL-based method shows robustness to these changes with similar performance metrics.

# 3. Discussion

To achieve high reliability and meet strict safety standards during real-world conditions, redundancy in sensing modalities is usually required. This requirement, coupled with typical operational conditions, brings about the need for frequent multi-sensor calibrations. Methods requiring specialized equipment and manual operation do not pose a viable and scalable solution, giving rise to the need for accurate, reliable, and automated methods capable of performing multi-sensor calibration in uncontrolled settings.

We take a holistic approach, also considering the system-level components required for scalable real-world deployment and propose two different approaches for calibration. The first is based on an optimization problem formulation of the calibration assignment, while the second is based on framing the task as a learning problem.

We base both proposed methods on a similar set of constraints that take account of the physical properties of the different signals as well as semantic information extracted from all sensing modalities, to produce pairwise alignment measures. Encapsulating these is the requirement that all cross-sensor transformations are globally self-consistent, meaning that a cyclic transformation should project data points to their original locations.

When viewing the calibration process as a component aimed for scalable and robust deployment there are considerable differences between the proposed optimization-based and SSL-based methods. The optimization-based method requires additional sub-components preceding the actual calibration in the form of procedures to aggregate appropriate samples for calibration and continuously querying some criteria for triggering re-calibration. These are important components which for the most part have previously been disregarded and are directly tied to the overall calibration performance. It is therefore our belief that they should be inseparable from the calibration method.

The SSL-based method requires a larger data collection effort prior to deployment in comparison to the optimization-based solution. Unlike most previous work, our proposed SSL-based method does not require any manual labelling or ground truth system. Instead, by using the set of constraints in the loss function during training, it is possible to train end-to-end in a self-supervised manner. Thus, greatly reducing the overall cost and resources required.

By visually comparing the two proposed methods, as shown in Fig. 3, it is evident that both methods can successfully solve the calibration task. Their desirability for use in specific applications is based on other system-level considerations, such as available computing resources during operation that might tip the scale in favor of the optimization-based method for example.

The results provided in Table 1 show that, by large, the SSL-based method achieves superior results to the optimization-based method. This implies that a DNN can potentially learn to extract and match features beyond the capability of its heuristics-based counterpart, allowing it to maintain its performance across varying scenes.

Additionally, we demonstrate the importance of imposing the global self-consistency constraint which yielded improved performance, as reflected by the overall lower error metrics. The results reported in Table 1 also suggest that apart from ensuring an overall self-consistent calibration, including the global constraint in the calibration process facilitates mutual corrective behavior between the different sensor pairs. Meaning, parameters derived reliably in one sensor pair can increase the reliability of parameters in other sensor pairs that now become coupled via the global constraint.

As previously mentioned, the optimization-based method includes components for sample aggregation and calibration triggering logics. The sample aggregation provides a variety of features across multiple frames, thereby increasing the optimization numerical stability. However, as the frame selection criteria refer to the presence or absence of certain features in the environment, in some cases the time required to gather enough samples might be significant. Similarly, calibration is triggered by misalignment measured with respect to particular features in the scene. If none can be found, then recalibration might be delayed.

In contrast, during deployment, the DNN can generate a calibration solution on a single frame basis. Meaning, there is no need for additional sample collection procedure or re-calibration triggering mechanism. This greatly simplifies the overall system design. In addition, this capability also makes the SSL-based method more robust and resilient to abrupt changes and sensor displacements that might occur in real-life settings due to vibrations, shock, temperature differences, and more. This is demonstrated in Fig. 4 where the differences between the uncalibrated and calibrated frames are clearly visible.

Driving environments and conditions are important considerations during real-world deployment. Since we aim to fulfil the desire for an automated, online calibration based solely on information gathered from the scene, we examined the effects of urban versus highway driving. As can be seen in extended data figures 1 and extended data figure 2, and quantitatively supported in extended data table 1, the SSL-method is more resilient to such changes with similar performance in both environments.

# 4. Methods

## 4.1. Physics and semantics informed calibration

A core concept of our proposed methods is to combine prior knowledge on the physical properties of different sensing modalities together with semantic information. The combination of the two is then used as constraints for cross-sensor alignment.

In this work we propose a holistic approach to the calibration problem in real-world conditions. We present two different calibration methods, the first is optimization-based and the second is SSL-based. Since these two are fundamentally different, the system-level architectures required for their deployment in a real-world setting vary significantly.

## 4.2. Feature extraction

Camera-semantics were extracted in the form of object detection and semantic segmentation using YOLOR[63] and SegFormer[64] respectively. Lidar point clouds underwent uniform sampling, which was previously found to benefit different feature extraction algorithms[65], and were then clustered using DBSCAN[66] to extract geometric objects in the scene. Lidar semantics utilized RangeNet++[67] and RANSAC[68] successively to segment the drivable area. Radar data underwent clustering similar to the lidar and tracking to distinguish moving objects. In addition, a weakly-supervised method[69] for creating a radar-based DNN was used to segment the drivable area.

## 4.3. Pairwise constraints

Three pairwise feature-based constraints were devised for each sensor pair that were a combination of physical properties of the sensing modalities and semantic information. These were later represented by smooth loss functions to be minimized in each of the methods.

With respect to the camera-lidar pairing, the projected lidar clusters are expected to correspond to relevant segmented objects in the image which are known to reflect lidar signals, such as: vehicles, pedestrians, signs, poles, buildings, fences, and vegetation. This was expressed as a loss term equal to the summed distances between the clusters and the segmented objects. As the sky is a known non-reflective region, lidar detections projected on regions in the image which were segmented as ‘sky’ or outside of the FOV were given a penalty.

The camera-radar pair utilized one of the radar's strong suites, which is the ability to measure Doppler. Since radar-based tracked clusters were necessarily associated with dynamic road users (vehicles, pedestrians, etc.), the associated loss was based on a sum of the distances between each projected radar-based tracked cluster and its nearest detected road user's center in the image. In addition, similarly to the camera-lidar projection, radar detections should not be projected on regions in the image which were segmented as 'sky' or outside of the camera FOV, hence any such occurrences were given a penalty.

The lidar-radar correspondence was reflected by the alignment between the projected lidar-based 'drivable area' and the radar-based 'drivable area' in the frame, measured by their IOU. In addition, since objects clustered in the lidar point cloud included dynamic road users, radar-based tracked clusters were expected to align with certain lidar clusters. The respective loss was defined as the sum of the distances between each projected radar-based tracked cluster and its nearest object cluster center in the lidar point cloud.

## 4.4. Global constraints

Under the assumption that the sensors are intrinsically calibrated, the transformations, now an expression of the extrinsic parameters alone, become linear and can be expressed in the $[\mathbf{R}|\mathbf{t}]$ matrix form of $\mathbf{T}_i^j = \begin{bmatrix} \mathbf{R}_i^j & \mathbf{t}_i^j \\ \mathbf{0}_{1\times 3} & 1 \end{bmatrix}$ where $\mathbf{R}_i^j$ and $\mathbf{t}_i^j$ are, respectively, the $3 \times 3$ rotation matrix and $3 \times 1$ translation vector from reference frame $i$ to $j$. By creating a closed-loop cyclic transformation, the global consistency constraint takes on the form of $\mathbf{T}_{lidar}^{camera} \cdot \mathbf{T}_{radar}^{lidar} \cdot \mathbf{T}_{camera}^{radar} = \mathbf{I}_{4\times 4}$ with $\mathbf{I}_{4\times 4}$ being a $4 \times 4$ identity matrix. For example, a point starting at some pixel on the camera, and undergoes transformation to the lidar, then the radar and back to camera, should return to its original pixel location. Similarly, with a lidar point or radar detection. In our implementation, as with the pairwise conditions, we regarded this constraint as an additional loss term to be minimized. We chose to express this loss as an explicit function of the cyclic transform's Euler angles and translation vector, equal to the $l_1$-norm of these elements.

## 4.5. Optimization-based method

The optimization-based calibration process included logics for continuous frame aggregation, recalibration triggers and calibration procedure, as illustrated in Fig. 5a. The input to the system is a set of frames comprised of pairwise, time-synced data from all 3 sensors (camera, lidar and radar). Semantic features are extracted from all frames, which are passed along with the original frames to a three-stage pairwise calibration pipeline. The pipeline filters the frames and uses the selected frames to detect miscalibration and perform pairwise optimization. Finally, the obtained pairwise transformations are optimized with respect to a global consistency measure to obtain the refined 3D transformations.

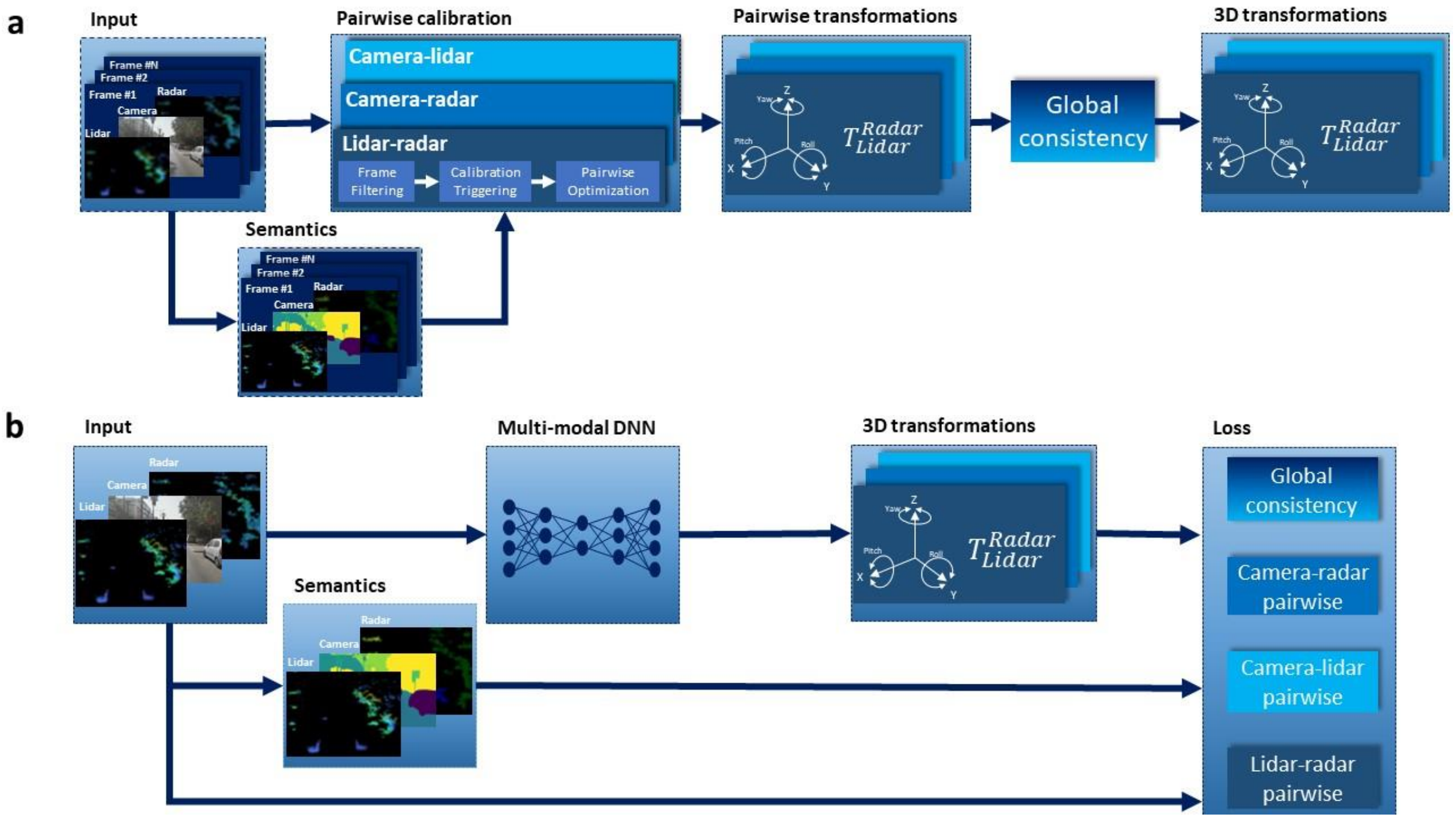

**Fig. 5 | Block diagram of the calibration methods**. **a**, Optimization-based method. The input to the system is a set of frames comprised of time-synced data from all three sensors (camera, lidar, radar). Semantic features are extracted from all frames, which are passed along with the original frames to a three-stage pairwise calibration pipeline. The pipeline filters the frames, uses the selected frames to detect miscalibration and perform pairwise optimization. Finally, the obtained pairwise transformations are optimized with respect to a global consistency measure to obtain the refined 3D transformations. **b**, SSL-based training method. The input is similar to the one in the optimization-based method, except that this method is a single frame method. The frame is passed on to a multi-modal DNN which simultaneously predicts 3D transformations for all sensor pairs, and to sensor specific DNNs which generate scene semantics. The output of all DNNs, along with the original frame, are all used to compute the various losses (pairwise losses and global self-consistency loss).

### 4.5.1. Frame aggregation and filtering

Samples corresponding to measurements from sensor pairs are collected continuously throughout the drive. Each of the two modalities' measurements undergoes filtering to determine whether the frame is suitable for calibration, and if so, both measurements are processed further to obtain distinctive features to be used in the calibration process of the respective modalities.

A unique set of requirements for a frame to be valid pertain to each modality pair: camera-lidar, camera-radar and lidar-radar. These criteria referred to the physics, semantics, and geometric conditions to be utilized in the calibration stage and were empirically found to encourage the numerical stability of the optimization process.

The conditions found in this study were occurrence of the semantic regions of 'sky' and 'drivable area' in the camera image of at least 20% and 10% respectively. In addition, a requirement of at least three instances of road users was also included. Lastly, we required the radar tracker to report at least three dynamic clusters. In all cases a time sync of under 5 msec between measurement pairs was required to alleviate temporal influence.

#### 4.5.2. Re-calibration triggers

The conditions for triggering the optimization-based re-calibration process are checked periodically with respect to the current set of gathered frames. These consist of three metrics, corresponding to each of the modalities pairs. Essentially, re-calibration between two sensing modalities is triggered when a significant misalignment between their measurements is detected. This is done based on the perceived association between different objects and/or regions identified in each of the sensors' measurements.

Misalignment for the camera-lidar pair is measured by summation of the number of projected lidar points that are projected outside of their associated segmentation in the image. Re-calibration is triggered if that sum exceeds 1% of the entire point cloud.

Similarly, misalignment for the camera-radar pair is measured by summation of the number of projected radar detections that fall outside of their associated segmentation in the image. Re-calibration is triggered if the number of dynamic detections outside of their associated object segmentation in the image is greater than 1.

Lastly, misalignment for the lidar-radar pair is measured by summation of the number of projected radar-based dynamic detections that fall outside of their associated objects in the lidar point cloud and by comparing semantic regions. Re-calibration is triggered if the number of dynamic detections projected outside of their respective lidar-based bounding boxes is greater than 1. Re-calibration could also be triggered if the percentage of lidar points is greater than 1% for those which are classified as 'drivable area' and are projected outside of the radar-based 'drivable area'.

#### 4.5.3. Optimization process

We formulated the calibration task as an optimization problem in which the following objective is to be minimized:

$$\min_{\mathbf{T}} L\big(L_P(\mathbf{T},\boldsymbol{\varphi}), L_G(\mathbf{T})\big) \tag{1}$$

with

$$L\big(L_P(\mathbf{T},\boldsymbol{\varphi}), L_G(\mathbf{T})\big) = [1 + L_P(\mathbf{T},\boldsymbol{\varphi})][1 + L_G(\mathbf{T})] \tag{2}$$

where $L_P$ and $L_G$ are the pairwise feature-based loss and global self-consistency loss respectively. $\mathbf{T}$ is the set of all pairwise rigid body transformations and $\boldsymbol{\varphi}$ is the set of all features across a given collection of frames for all modalities pairs.

The optimization was run with respect to triggered pairs of modalities as well as the entire sensor suite. The loss was accumulated only over samples from uncalibrated sensor couples with the addition of the global self-consistency term. The solution was obtained iteratively using the sample-based Differential Evolution[70] and Nelder-Mead[71] algorithms consecutively.

## 4.2. SSL-based method

The SSL-based method is fundamentally different from the optimization-based method in several aspects affecting its design and deployment. The SSL-based method aims to learn an adaptive response to a set of camera-lidar-radar trio. Meaning, during inference the DNN does not require any additional semantic information and can be applied on a single-frame basis. As a results, there is no need for real time sample collection or triggering mechanisms, making the overall system design much simpler and robust.

### 4.2.1. Training methodology

The SSL-based method utilizes end-to-end training where the DNN take as input a trio of uncalibrated RGB image, lidar point cloud and radar point cloud and outputs a set of 3D transformation matrices for each of the sensing modalities pairs. The DNN is trained in a self-supervised manner and no labels or ground truth information is used.

During training, the output of the DNN, together with semantic information extracted from the camera-lidar-radar trio and the measurements themselves are used to calculate the pairwise losses as well as the global self-consistency loss as illustrated in Fig 5b.

The training dataset consists of multiple data-collection sequences, each with different sensors alignment making the dataset varied and more challenging. To augment the variability and improve the generalization of our SSL-based method, we randomly rotated and translated the lidar and radar point clouds during training to simulate a much larger set of uncalibrated situations.

### 4.2.2. Model

The model is illustrated in Fig. 6 and consists of three sensor-wise encoders, a global fusion module and separate pairwise heads. The three encoders extract separate features for each modality, these can be implemented as convolutional layers, transformers, or multi-layer perceptron. These features are then concatenated and fed into a global fusion module, which regresses the transformation parameters between all sensor pairs. The information then flows to three pairwise heads each outputting the corresponding 6 degrees of freedom represented by three Euler angles and three translation components in 3D cartesian space. Similarly, to the encoders, the fusion module and the pairwise heads can be implemented using a variety of flavours. In our implementation we chose a convolutional approach for simplicity.

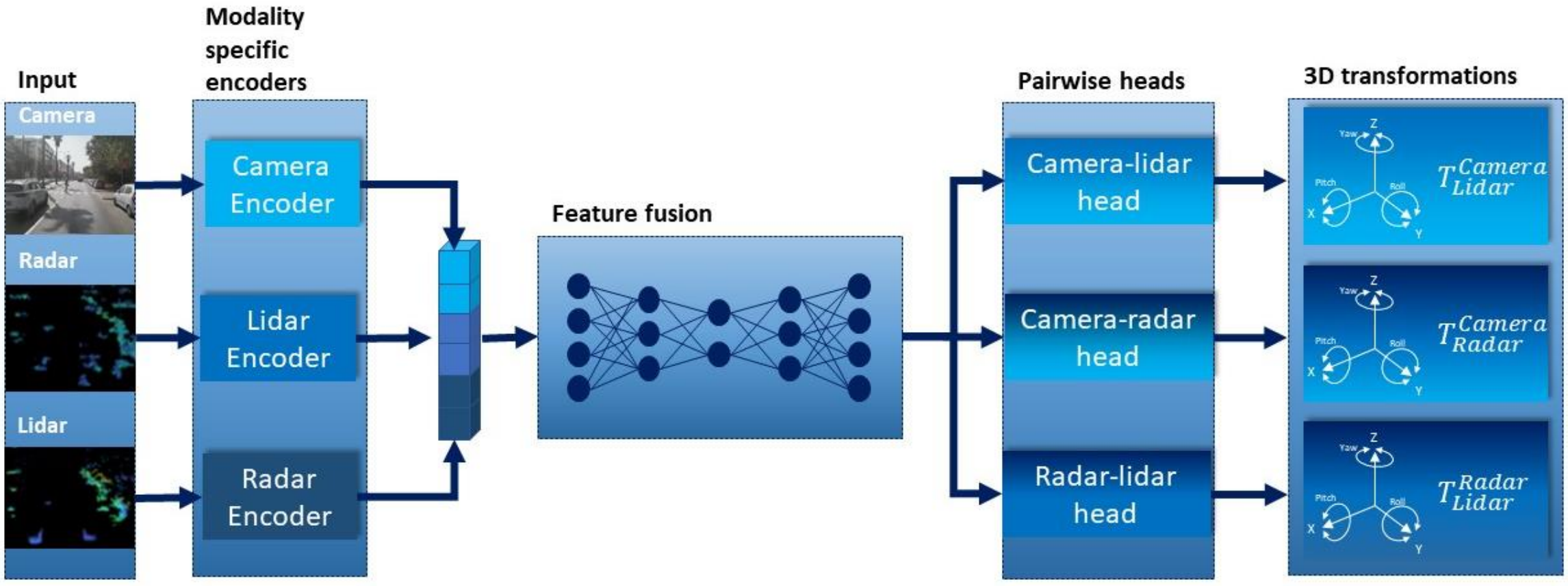

**Fig. 6 | Multi-modal DNN architecture.** The network encodes sensor data per modality, concatenates the encoded features and feeds them into a feature fusion network. To generate the required pairwise transformations, the network splits outputs to three heads, one for each sensor pair.

### 4.2.3. Loss function

The loss function for the SSL-based method is a sum of the pairwise and global self-consistency loss: $L = L_P(\mathbf{T_\theta}(\boldsymbol{x}), \boldsymbol{\varphi}_\mathbf{x}) + L_G(\mathbf{T_\theta}(\boldsymbol{x}))$. With $\mathbf{T_\theta}(\boldsymbol{x})$ being the set of all pairwise rigid body transformations learned by the DNN which is parametrized by the weights $\boldsymbol{\theta}$ on frame $\boldsymbol{x}$; $\boldsymbol{\varphi}_\mathbf{x}$ is the set of all precomputed semantic and physical features for all sensing modalities of frame $\boldsymbol{x}$.

### 4.2.4. Implementation details

Training was implemented in PyTorch, the optimizer used was Adam with $\beta_1 = 0.9$, $\beta_2 = 0.999$, batch size was 4, and learning rate used cosine decay from $3 \cdot 10^{-4}$ to $3 \cdot 10^{-7}$. All configurations were trained until convergence on a single A6000 GPU which took about 20 epochs.

# 5. Conclusion

In this work we address the critical issue of online multi-sensor calibration in uncontrolled environments. To that end we consider a setup including the currently most prominent sensors for automotive perception purposes: camera, lidar, and radar. We propose a physics and semantic informed methodology by which cross-sensor alignments are gauged based on matching semantics and conformation to known physical properties of the different sensing modalities.

We propose two fundamentally different approaches to the calibration problem: an optimization-based method and an SSL-based method. Under an optimization framework we provide a representative minimization problem in which a calculable measure of pairwise misalignments and global self-inconsistency is to be minimized. The second approach is in the form of a DNN trained in a self-supervised manner. Meaning, this method does not rely on any form of ground-truth data.

The optimization-based method, empirically shown to require the stabilizing effect of running on multiple measurement frames, is accompanied by frame aggregation and filtering logics. Furthermore, considering real-world application, recalibration triggering is conceived and included as well. Conversely, the SSL-based solution can be self-contained, as a result of its demonstrated capacity for precise calibration from a single frame. Both solutions are shown to provide precise and robust joint calibration in a real-world setting, beyond the capabilities of current state-of-the-art methods. Experimental results support our selection of semantic features and characteristics of the different sensing modalities. Our inclusion of the self-consistency constraint in both approaches is shown to significantly improve the calibration results, as a facilitator of mutual corrective behavior between the different sensor pairs.

In a broader sense, this work shows an example case of how direct-optimization and self-supervision are closely related, each with its own pros and cons depending on the task and use case. We hope this work will inspire additional research in self-supervised learning and open new avenues traditionally only solved by optimization frameworks.

# References

1. Yeong, D. J., Velasco-hernandez, G., Barry, J. & Walsh, J. Sensor and sensor fusion technology in autonomous vehicles: A review. *Sensors* vol. 21 2140 (2021).

2. Wei, Z. *et al.* MmWave radar and vision fusion for object detection in autonomous driving: A review. *Sensors* **22**, 2542 (2021).

3. Cui, Y. *et al.* Deep learning for image and point cloud fusion in autonomous driving: A review. *IEEE Transactions on Intelligent Transportation Systems* vol. 23 722–739 (2022).

4. Lekic, V. & Babic, Z. Automotive radar and camera fusion using generative adversarial networks. *Computer Vision and Image Understanding* **184**, 1–8 (2019).

5. Haag, S. *et al.* OAFuser: Online adaptive extended object tracking and fusion using automotive radar detections. in *IEEE International Conference on Multisensor Fusion and Integration for Intelligent Systems (MFI)* 303–309 (2020).

6. Chen, C. *et al.* Selective sensor fusion for neural visual-inertial odometry. in *IEEE/CVF Conference on Computer Vision and Pattern Recognition* 10542–10551 (2019).

7. Mirzaei, F. M., Kottas, D. G. & Roumeliotis, S. I. 3D lidar-camera intrinsic and extrinsic calibration: Identifiability and analytical least-squares-based initialization. *International Journal of Robotics Research* **31**, 452–467 (2012).

8. Zhou, L., Li, Z. & Kaess, M. Automatic extrinsic calibration of a camera and a 3D lidar using line and plane correspondences. in *IEEE/RSJ International Conference on Intelligent Robots and Systems (IROS)* 5562–5569 (IEEE, 2018).

9. Owens, J. L., Osteen, P. R. & Daniilidis, K. MSG-Cal: Multi-sensor graph-based calibration. in *IEEE/RSJ International Conference on Intelligent Robots and Systems (IROS)* 3660–3667 (2015).

10. Yin, L. *et al.* CoMask: Corresponding mask-based end-to-end extrinsic calibration of the camera and lidar. *Remote Sensing* **12**, 1925 (2020).

11. Huang, J. K. & Grizzle, J. W. Improvements to target-based 3D lidar to camera calibration. *IEEE Access* **8**, 134101–134110 (2020).

12. Peršić, J., Marković, I. & Petrović, I. Extrinsic 6DoF calibration of a radar–lidar–camera system enhanced by radar cross section estimates evaluation. *Robotics and Autonomous Systems* **114**, 217–230 (2019).

13. Peršić, J., Marković, I. & Petrović, I. Extrinsic 6DoF calibration of 3D lidar and radar. in *European Conference on Mobile Robots (ECMR)* 1–6 (2017).

14. Domhof, J., Kooij, J. F. P. & Gavrila, D. M. A joint extrinsic calibration tool for radar, camera and lidar. *IEEE Transactions on Intelligent Vehicles* **6**, 571–582 (2021).

15. Oh Jiyong, Kim Ki-Seok, Park Miryong & Kim Sungho. A comparative study on camera-radar calibration methods. in *15th International Conference on Control, Automation, Robotics and Vision (ICARCV)* 1057–1062 (IEEE, 2018).

16. Jiang, J. *et al.* Line feature based extrinsic calibration of lidar and camera. in *2018 IEEE International Conference on Vehicular Electronics and Safety, ICVES 2018* (Institute of Electrical and Electronics Engineers Inc., 2018). doi:10.1109/ICVES.2018.8519493.

17. Munoz-Banon, M. A., Candelas, F. A. & Torres, F. Targetless camera-lidar calibration in unstructured environments. *IEEE Access* **8**, 143692–143705 (2020).

18. Yuan, C., Liu, X., Hong, X. & Zhang, F. Pixel-level extrinsic self calibration of high resolution lidar and camera in targetless environments. *IEEE Robotics and Automation Letters* **6**, 7517–7524 (2021).

19. Tamas, L. & Kato, Z. Targetless calibration of a lidar-perspective camera pair. in *IEEE International Conference on Computer Vision* 668–675 (Institute of Electrical and Electronics Engineers Inc., 2013). doi:10.1109/ICCVW.2013.92.

20. Nagy, B. & Benedek, C. On-the-fly camera and lidar calibration. *Remote Sensing* **12**, (2020).

21. Heng, L. Automatic targetless extrinsic calibration of multiple 3D lidars and radars. in *IEEE International Conference on Intelligent Robots and Systems* 10669–10675 (Institute of Electrical and Electronics Engineers Inc., 2020). doi:10.1109/IROS45743.2020.9340866.

22. Chien, H. J., Klette, R., Schneider, N. & Franke, U. Visual odometry driven online calibration for monocular lidar-camera systems. in *International Conference on Pattern Recognition* vol. 0 2848–2853 (Institute of Electrical and Electronics Engineers Inc., 2016).

23. Horn, M., Wodtko, T., Buchholz, M. & Dietmayer, K. Online extrinsic calibration based on per-sensor ego-motion using dual quaternions. *IEEE Robotics and Automation Letters* **6**, 982–989 (2021).

24. Wise, E., Persic, J., Grebe, C., Petrovic, I. & Kelly, J. A continuous-time approach for 3D radar-to-camera extrinsic calibration. in *IEEE International Conference on Robotics and Automation (ICRA)* 13164–13170 (Institute of Electrical and Electronics Engineers (IEEE), 2021). doi:10.1109/icra48506.2021.9561938.

25. Peršić, J., Petrović, L., Marković, I. & Petrović, I. Spatio-temporal multisensor calibration based on gaussian processes moving object tracking. in *arXiv preprint arXiv:1904.04187* (2019).

26. Napier, A., Corke, P. & Newman, P. Cross-calibration of push-broom 2D lidars and cameras in natural scenes. in *Proceedings - IEEE International Conference on Robotics and Automation* 3679–3684 (2013). doi:10.1109/ICRA.2013.6631094.

27. Miled, M., Soheilian, B., Habets, E. & Vallet, B. Hybrid online mobile laser scanner calibration through image alignment by mutual information. *ISPRS Annals of Photogrammetry, Remote Sensing and Spatial Information Sciences* **III–1**, 25–31 (2016).

28. Pandey, G., McBride, J. R., Savarese, S. & Eustice, R. M. Automatic extrinsic calibration of vision and lidar by maximizing mutual information. *Journal of Field Robotics* **32**, 696–722 (2015).

29. Taylor, Z. & Nieto, J. A mutual information approach to automatic calibration of camera and lidar in natural environments. in *Australian Conference on Robotics and Automation* 3–5 (2012).

30. Jiang, P., Osteen, P. & Saripalli, S. SemCal: Semantic lidar-camera calibration using neural mutual information estimator. in *IEEE International Conference on Multisensor Fusion and Integration for Intelligent Systems (MFI)* 1–7 (Institute of Electrical and Electronics Engineers Inc., 2021). doi:10.1109/MFI52462.2021.9591203.

31. Liu, Z., Tang, H., Zhu, S. & Han, S. SemAlign: Annotation-free camera-lidar calibration with semantic alignment loss. in *IEEE/RSJ International Conference on Intelligent Robots and Systems (IROS)* 8845–8851 (2021).

32. Zhu, Y., Li, C. & Zhang, Y. Online camera-lidar calibration with sensor semantic information. in *IEEE International Conference on Robotics and Automation (ICRA)* 4970–4976 (2020).

33. Wang, W., Nobuhara, S., Nakamura, R. & Sakurada, K. SOIC: Semantic Online Initialization and Calibration for LiDAR and Camera. (2020).

34. Schneider, N., Piewak, F., Stiller, C. & Franke, U. RegNet: Multimodal sensor registration using deep neural networks. in *IEEE Intelligent Vehicles Symposium, Proceedings* 1803–1810 (Institute of Electrical and Electronics Engineers Inc., 2017). doi:10.1109/IVS.2017.7995968.

35. Iyer, G., Ram, R. K., Murthy, J. K. & Krishna, K. M. CalibNet: Geometrically supervised extrinsic calibration using 3D spatial transformer networks. in *IEEE/RSJ Intelligent Robots and Systems (IROS)* 1110–1117 (IEEE, 2018).

36. Shi, J. *et al.* CalibRCNN: Calibrating Camera and LiDAR by recurrent convolutional neural network and geometric constraints. in *IEEE International Conference on Intelligent Robots and Systems* 10197–10202 (Institute of Electrical and Electronics Engineers Inc., 2020). doi:10.1109/IROS45743.2020.9341147.

37. Yuan, K., Guo, Z. & Jane Wang, Z. RGGNet: Tolerance aware lidar-camera online calibration with geometric deep learning and generative model. *IEEE Robotics and Automation Letters* **5**, 6956–6963 (2020).

38. Lv, X., Wang, B., Dou, Z., Ye, D. & Wang, S. LCCNet: Lidar and camera self-calibration using cost volume network. in *IEEE/CVF Conference on Computer Vision and Pattern Recognition* 2894–2901 (2021).

39. Lv, X., Wang, S. & Ye, D. CFNet: Lidar-camera registration using calibration flow network. *Sensors* **21**, 8112 (2021).

40. Jing, X., Ding, X., Xiong, R., Deng, H. & Wang, Y. DXQ-Net: Differentiable LiDAR-Camera Extrinsic Calibration Using Quality-aware Flow. in *arXiv Preprint* (2022).

41. Schneider, N., Piewak, F., Stiller, C. & Franke, U. RegNet: Multimodal sensor registration using deep neural networks. in *IEEE Intelligent Vehicles Symposium, Proceedings* 1803–1810 (Institute of Electrical and Electronics Engineers Inc., 2017). doi:10.1109/IVS.2017.7995968.

42. Orr, I. *et al.* Coherent, super-resolved radar beamforming using self-supervised learning. *Science Robotics* **6**, 431 (2021).

43. Laine, S., Karras, T., Lehtinen, J. & Aila, T. High-quality self-supervised deep image denoising. *Advances in Neural Information Processing Systems* **32**, (2019).

44. Zhang, Y., Tian, Y., Ukong, Y., Zhong, B. & Fu, Y. Residual dense network for image super-resolution. in *IEEE conference on computer vision and pattern recognition* 2472–2481 (2018).

45. Wang, Z., Chen, J. & Hoi, S. C. H. Deep Learning for Image Super-Resolution: A Survey. *IEEE Transactions on Pattern Analysis and Machine Intelligence* vol. 43 3365–3387 (2021).

46. Dong, C., Loy, C. C., He, K. & Tang, X. Image Super-Resolution Using Deep Convolutional Networks. *IEEE Transactions on Pattern Analysis and Machine Intelligence* **38**, 295–307 (2016).

47. Qian, Y., Hou, J., Kwong, S. & He, Y. PUGeo-Net: A geometry-centric network for 3D point cloud upsampling. in *European Conference on Computer Vision* 752–769 (2020).

48. Li, R., Li, X., Fu, C.-W., Cohen-Or, D. & Heng, P.-A. Pu-gan: a point cloud upsampling adversarial network. in *IEEE/CVF International Conference on Computer Vision* 7203–7212 (2019).

49. Qian, G., Abualshour, A., Li, G., Thabet, A. & Ghanem, B. Pu-gcn: Point cloud upsampling using graph convolutional networks. in *IEEE/CVF Conference on Computer Vision and Pattern Recognition* 11683–11692 (2021).

50. Beker, D. *et al.* Monocular differentiable rendering for self-supervised 3D object detection. in *European Conference on Computer Vision* 514–529 (2020).

51. Lee, W., Na, J. & Kim, G. Multi-task self-supervised object detection via recycling of bounding box annotations. in *IEEE/CVF Conference on Computer Vision and Pattern Recognition* 4984–4993 (2019).

52. Karthik Mustikovela, S. *et al.* Self-supervised object detection via generative image synthesis. in *IEEE/CVF International Conference on Computer Vision* 8609–8618 (2021).

53. Zhan, X., Liu, Z., Luo, P., Tang, X. & Loy, C. C. Mix-and-match tuning for self-supervised semantic segmentation. *Proceedings of the AAAI Conference on Artificial Intelligence* **32**, (2018).

54. Wang, Y., Zhang, J., Kan, M., Shan, S. & Chen, X. Self-supervised equivariant attention mechanism for weakly supervised semantic segmentation. in *IEEE/CVF Conference on Computer Vision and Pattern Recognition* 12275–12284 (2020).

55. Araslanov, N. & Roth, S. Self-supervised augmentation consistency for adapting semantic segmentation. in *IEEE/CVF Conference on Computer Vision and Pattern Recognition* 15384–15394 (2021).

56. Wang, Y. & Solomon, J. PRNet: Self-supervised learning for partial-to-partial registration. *Adv Neural Inf Process Syst* **32**, (2019).

57. Li, H. & Fan, Y. Non-rigid image registration using self-supervised fully convolutional networks without training data. in *Proceedings - International Symposium on Biomedical Imaging* vols. 2018-April 1075–1078 (IEEE Computer Society, 2018).

58. Yang, H., Dong, W. & Koltun, V. Self-supervised geometric perception. in *IEEE/CVF Conference on Computer Vision and Pattern Recognition* 14350–14361 (2021).

59. Fang, J. *et al.* Self-supervised camera self-calibration from video. in *IEEE International Conference on Robotics and Automation (ICRA)* (2021).

60. Peršić, J., Petrović, L., Marković, I. & Petrović, I. Online multi-sensor calibration based on moving object tracking. *Advanced Robotics* **35**, 130–140 (2021).

61. Wang, T., Zhu, X., Pang, J. & Lin, D. FCOS3D: Fully convolutional one-stage monocular 3D object detection. in *IEEE/CVF International Conference on Computer Vision* 913–922 (2021).

62. Duan, K. *et al.* CenterNet: Keypoint triplets for object detection. in *IEEE/CVF international conference on computer vision* 6569–6578 (2019).

63. Wang, C. Y., Yeh, I. H. & Liao, H. Y. M. You only learn one representation: Unified network for multiple tasks. in *arXiv preprint arXiv:2105.04206* (2021).

64. Xie, E. *et al.* SegFormer: Simple and efficient design for semantic segmentation with transformers. *Advances in Neural Information Processing Systems* **34**, (2021).

65. Orr, I. *et al.* Effects of lidar and radar resolution on DNN-based vehicle detection. *Journal of the Optical Society of America A* **38**, B29 (2021).

66. Ester, M., Kriegel, H.-P., Sander, J. & Xu, X. A density-based algorithm for discovering clusters in large spatial databases with noise. *KDD* **96**, 226–231 (1996).

67. Milioto, A., Vizzo, I., Behley, J. & Stachniss, C. RangeNet++: Fast and accurate lidar semantic segmentation. in *IEEE/RSJ International Conference on Intelligent Robots and Systems (IROS)* 4213–4220 (2019).

68. Foley, J. D., Fischler, M. A. & Bolles, R. C. Random sample consensus: A paradigm for model fitting with applications to image analysis and automated cartography. *Commun ACM* **24**, 381–395 (1981).

69. Orr, I., Cohen, M. & Zalevsky, Z. High-resolution radar road segmentation using weakly supervised learning. *Nature Machine Intelligence* **3**, 239–246 (2021).

70. Storn, R. & Price, K. Differential evolution–a simple and efficient heuristic for global optimization over continuous spaces. *Journal of Global Optimization* **11**, 341–359 (1997).

71. Nelder, J. A. & Mead, R. A simplex method for function minimization. *Comput J* **7**, 308–313 (1965).

# Extended data

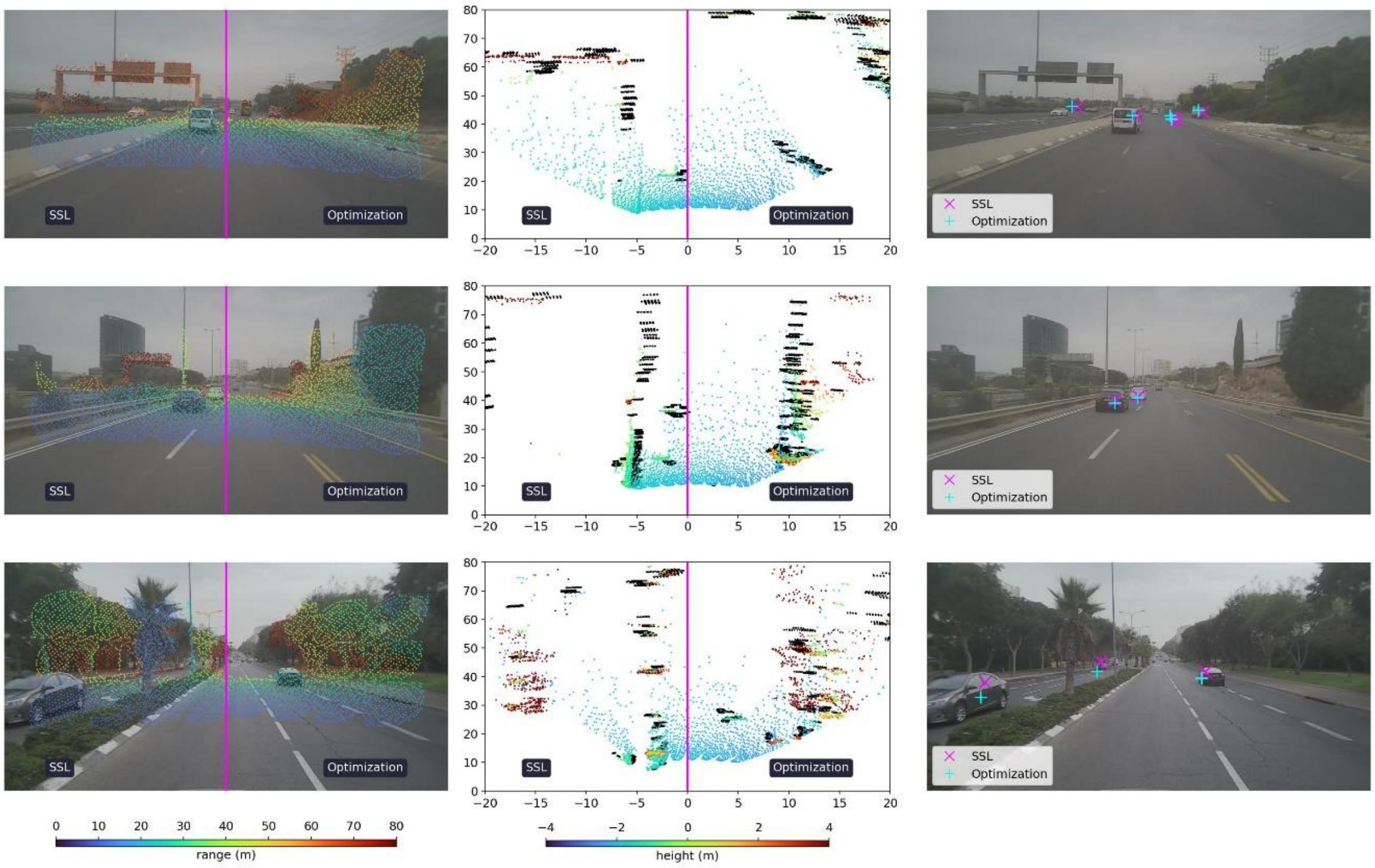

**Extended data figure 1 | Sample frames in highway environment.** From left to right: Calibration results of the optimization-based and SSL-based methods for camera-lidar, lidar-radar and camera-radar, respectively. On the left column, the camera-lidar calibration showing the projected lidar point cloud onto its corresponding image, with colormap representing range in meters. In the middle column, the lidar-radar calibration is portrayed in bird's eye view. The lidar is represented in a colormap for height in meters, whereas the radar detections are represented as black points. For both pairs (camera-lidar and lidar-radar), each frame is split along the middle, where the optimization-based calibration is displayed in the right half and the SSL-based calibration in the left half. On the right column, center of mass from tracked radar clusters are projected onto the corresponding images using the calibrations from both proposed methods.

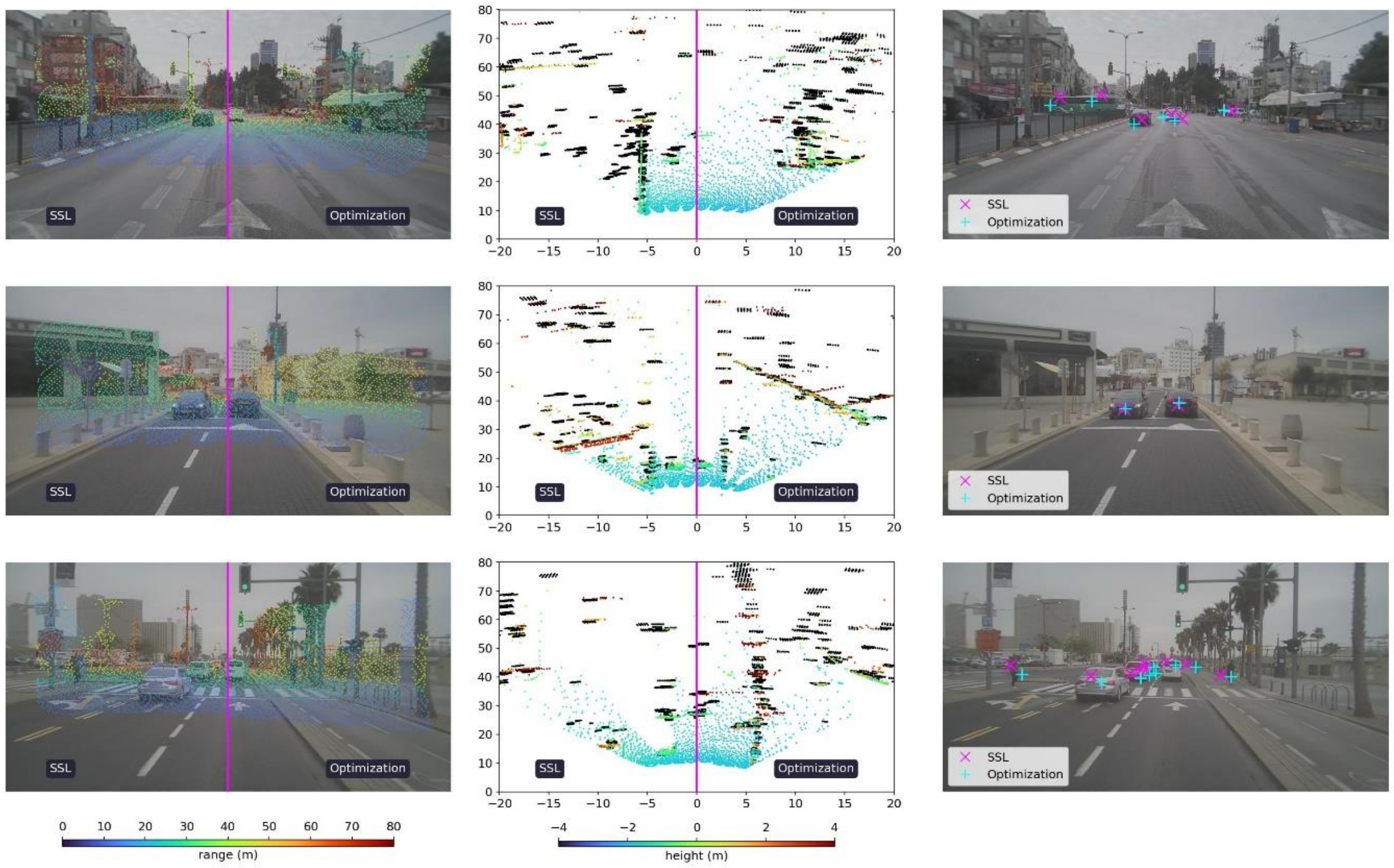

**Extended data figure 2 | Sample frames in urban environment.** From left to right: Calibration results of the optimization-based and SSL-based methods for camera-lidar, lidar-radar and camera-radar, respectively. On the left column, the camera-lidar calibration showing the projected lidar point cloud onto its corresponding image, with colormap representing range in meters. In the middle column, the lidar-radar calibration is portrayed in bird's eye view. The lidar is represented in a colormap for height in meters, whereas the radar detections are represented as black points. For both pairs (camera-lidar and lidar-radar), each frame is split along the middle, where the optimization-based calibration is displayed in the right half and the SSL-based calibration in the left half. On the right column, center of mass from tracked radar clusters are projected onto the corresponding images using the calibrations from both proposed methods.

| | | | **Pairwise calibration errors** | | | | | | | | | **Global calibration errors** | | |
|---|---|---|---|---|---|---|---|---|---|---|---|---|---|---|
| | | | **Camera-Lidar** | | | **Camera-Radar** | | | **Lidar-Radar** | | | **Camera-Lidar-Radar** | | |
| | | | **Az [deg]** | **El [deg]** | **R [m]** | **Az [deg]** | **El [deg]** | **R [m]** | **Az [deg]** | **El [deg]** | **R [m]** | **Az [deg]** | **El [deg]** | **R [m]** |
| **Environment** | **High-way** | Joint Optimization | 0.42 | 0.65 | - | 0.08 | 0.60 | - | 0.31 | 1.17 | 0.13 | 0.17 | 0.02 | 0.01 |
| | | Joint SSL | 0.01 | 0.22 | | 0.09 | 0.08 | | 0.49 | 0.66 | 0.14 | 0.06 | 0.03 | 0.07 |
| | **Urban** | Joint Optimization | 0.13 | 0.21 | - | 0.13 | 0.49 | - | 0.02 | 0.69 | 0.13 | 0.06 | 0.01 | 0.01 |
| | | Joint SSL | 0.10 | 0.18 | | 0.03 | 0.37 | | 0.51 | 0.61 | 0.13 | 0.03 | 0.04 | 0.04 |

**Extended data table 1 | Quantitative evaluation of the effects of driving environments.** Average calibration errors, computed with respect to three reflective corners, are provided for all modality pairs as well as for the closed transformation loop between all sensors. The optimization-based method performs better in urban environment than in highway environment. The SSL-based methods show robustness to these changes with similar performance metrics.